# Revisiting Deep Intrinsic Image Decompositions

Qingnan Fan[*1]   Jiaolong Yang[2]   Gang Hua[2]   Baoquan Chen[3,4]   David Wipf[2]
[1]Shandong University   [2]Microsoft Research
[3]Shenzhen Research Institute, Shandong University   [4]Peking University
fqnchina@gmail.com, {jiaoyan,ganghua,davidwip}@microsoft.com, baoquan@pku.edu.cn

## Abstract

*While invaluable for many computer vision applications, decomposing a natural image into intrinsic reflectance and shading layers represents a challenging, underdetermined inverse problem. As opposed to strict reliance on conventional optimization or filtering solutions with strong prior assumptions, deep learning based approaches have also been proposed to compute intrinsic image decompositions when granted access to sufficient labeled training data. The downside is that current data sources are quite limited, and broadly speaking fall into one of two categories: either dense fully-labeled images in synthetic/narrow settings, or weakly-labeled data from relatively diverse natural scenes. In contrast to many previous learning-based approaches, which are often tailored to the structure of a particular dataset (and may not work well on others), we adopt core network structures that universally reflect loose prior knowledge regarding the intrinsic image formation process and can be largely shared across datasets. We then apply flexibly supervised loss layers that are customized for each source of ground truth labels. The resulting deep architecture achieves state-of-the-art results on all of the major intrinsic image benchmarks, and runs considerably faster than most at test time.*

## 1. Introduction

The decomposing of natural images into multiple intrinsic layers can serve a variety of high-level vision tasks such as 3D object compositing, surface re-texturing, and relighting [3]. In this regard, the core intrinsic image model we consider here is predicated on an ideal diffuse environment, in which an input image $I$ is the pixel-wise product of an albedo or reflectance image $R$ and a shading image $S$, i.e.,

$$I \approx R \odot S. \quad (1)$$

The albedo layer indicates how object surface materials reflect light, while shading accounts for illumination effects due to geometry, shadows, and interreflections. While obviously useful, estimating such a decomposition is a fundamentally ill-posed problem as there exist infinitely many feasible solutions to (1). Fortunately though, prior information, often instantiated via specially tailored image smoothing filters or energy terms, allows us to constrain the space of feasible solutions [1, 2, 3, 10, 20, 23]. For example, the albedo image will usually be approximately piecewise constant, with a finite number of levels reflecting a discrete set of materials and boundaries common to natural scenes. In contrast, the shading image is often assumed to be greyscale, and is more likely to contain smooth gradations quantified by small directional derivatives except at locations with cast shadows or abrupt changes in scene geometry [16].

On the other hand, given access to ground truth intrinsic image decompositions, deep convolutional neural networks (CNN), at least in principle, provide a data-driven candidate for solving this ill-posed inverse problem with fewer potentially heuristic or hand-crafted assumptions. However, ground truth data that sufficiently covers the rich variety inherent to natural scenes, and includes dense intrinsic labels across entire images, is extremely difficult to acquire. Consequently, existing databases are each limited in various different ways, and thus far, state-of-the-art deep network models built using them likewise display a high degree of dataset-dependent architectural variance, *i.e.*, to achieve the best results, significantly different network architectures have been applied that compensate for each nuanced data source.

For instance, the MIT intrinsic dataset [14] is limited to images of single, specialized objects, which lacks diversity and scene-level realism for training a network that generalizes to broader scenarios. On the other hand, the MPI-Sintel benchmark is rendered on an open source animation movie [4]. Their rendered images often lack realism, and traditional deep networks trained on these data may perform poorly on more natural examples [22]. Finally then, to overcome the above downsides, the Intrinsic Images in the Wild (IIW) dataset was created from real-world photos

---

[*]This work was done when Qingnan Fan was an intern at MSR.

[2]. Although dense ground truth decompositions are not available, pairwise reflectance comparisons have been labeled via Amazon Mechanical Turk for a sparse collection of points in each image.

To summarize then, there presently exists a trade-off between realistic yet weakly supervised image sources (*e.g.*, IIW with sparse, pairwise comparison labels) and synthetic or highly-controlled sources blessed with dense ground truth labels (*e.g.*, MIT and MPI-Sintel). In general, previous learning-based solutions have invoked network designs and training pipelines specifically tailored for a particular data source. But if our ultimate goal is a model that can eventually transfer to practical environments, then it behooves us to consider data-set-independent architectures. Or stated conversely, if a different nuanced model structure is required to obtain state-of-the-art results on each different intrinsic image benchmark (all of which have significant shortcomings as mentioned above), then how confident can we be that any one such structure will effectively translate to broader application scenarios with more diverse input sources? For this reason we consider a quasi-universal architecture in the sense that, small differences in parameterizations to account for dataset size/type notwithstanding, the high-level pipeline itself is identical whether training is performed using samples formed from dense maps (MPI, MIT-Sintel) or pair-wise comparisons (IIW).

To accomplish this we allow flexible supervision layers to serve as an intermediary between diverse training data sources and an otherwise fixed network architecture. The latter is chosen to reflect basic universal assumptions describing intrinsic image decompositions independent of any one data particular set. For example, we assume that the albedo component is a priori likely to be piecewise constant or flattened, reflecting broad areas of identical reflectance and abrupt changes to new material surfaces. Such a prior should be broadly effective regardless of available supervision. We incorporate this knowledge via three network substructures: (i) a direct intrinsic network to predict a coarse first estimate of the albedo and/or shading image, (ii) an independent guidance network to predict the significant edges that largely originate from the albedo layer, and (iii) a 1D recursive domain filter that uses the output of the guidance network to steer the final albedo estimate towards a piecewise constant or flattened image. The entire process is differentiable and amenable to end-to-end training.

Our overall contributions can be summarized as follows:

- We provide the first demonstration of a single basic deep architecture capable of achieving state-of-the-art results when applied to each of the major intrinsic benchmarks, despite the radically different nature of the underlying data types. Unlike previous approaches, we accomplish this by modifying the training objective via flexible supervision layers without the need to significantly modify the overarching network structure itself, which is based on loose prior assumptions naturally satisfied by real images.

- On the most challenging IIW data, we provide the first trainable end-to-end system that can both produce state-of-the-art results on supervised pairwise comparison metrics computed from sparse points, while simultaneously generating a plausible, piecewise-flat dense map to characterize all other unsupervised image locations.

- We achieve significant improvements over both unseen indoor and outdoor real images via joint training of multiple data sources. We demonstrate that the well-known limitations of the existing dataset can be overcome by incorporating other types of training samples.

- We accomplish each of the above via a system requiring a minimal computational footprint at test time, with execution speeds comparable or considerably faster than existing alternatives.

## 2. Related Work

Deep learning has been widely used in many image processing tasks [6, 7, 8, 18, 24, 25] across various datasets, such as Intrinsic Images in the Wild (IIW). For example, [18] learns a local linear classifier using deep features and contextual clues present in two local image patches. Alternatively, in [24] a multi-stream network architecture is learned whose input source comes not only from the local surrounding patch of compared points, but also from the global image. Moreover, to estimate a globally consistent albedo layer, a second, relative reflectance classification step is incorporated via optimization of a hinge loss. Similarly, [25] also learns a deep network to classify the pairwise points from both local and global contextual information. Afterwards, they yield a piecewise constant albedo image by segmenting the input image into constant superpixels and optimizing a quadratic objective function.

Note that each of the above examples treat the intrinsic decomposition as a classification problem, and ultimately require feeding every pair of patches to the trained deep network to predict the relative reflectance of a new image, which is very computationally-intensive. In contrast, [19] attempts to first predict a dense reflectance layer via a convolutional neural network by supervising the sparse pairwise points of IIW using a similar hinge loss. Given that such a predicted image will not generally meet the piecewise constancy requirement of albedo layers, they execute a second post-processing step using [3] to flatten the dense map through a guided filter or joint bilateral filter.

Several existing deep network pipelines have also been built using the MPI-Sintel and MIT datasets with dense ground truth labels. First, [22] learns a two-scale convolutional network to directly predict both albedo and shading images. However, the specific architecture, which closely resembles that from [11] developed for predicting depth and surface normals, involves intermediate feature maps at 1/32 scale such that significant detail information may be compromised. A second more recent method from [21] trains an encoder-decoder CNN to learn albedo, shading and specular images with millions of object-level synthetic intrinsic images via rendering ShapeNet [5] [1]; however, this approach does not apply to scene-level images as we consider herein.

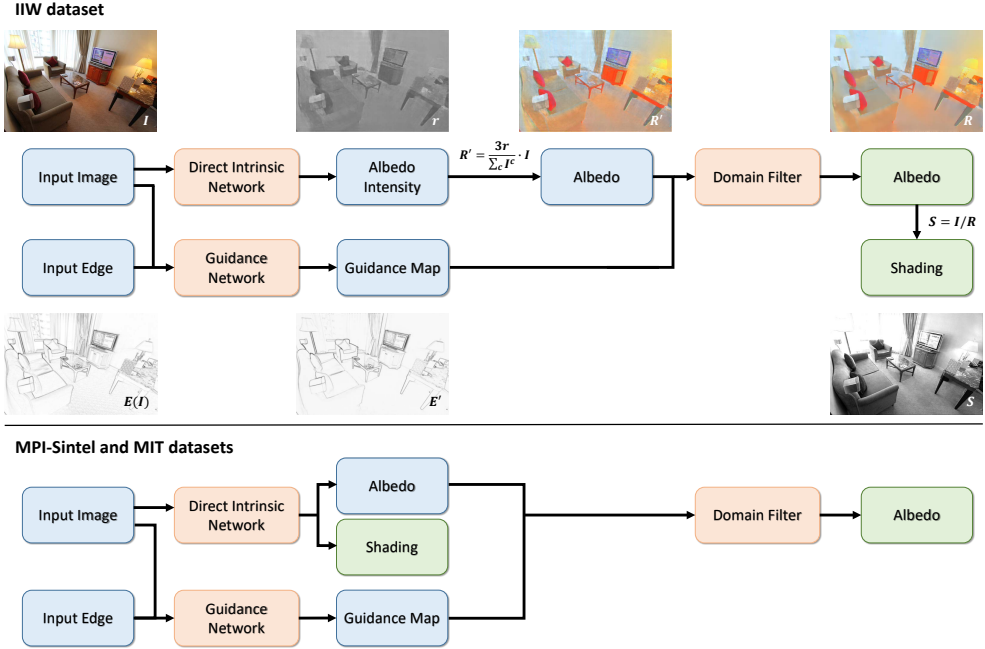

Figure 1. The proposed framework. Our end-to-end trainable intrinsic image estimation framework, which produces a flattened albedo image and a realistic shading image. Orange boxes indicate shared network structures, while green boxes represent the final network outputs and flexible loss layers that vary based on available supervision, either weak pairwise comparisons, e.g., IIW (*top*), or full dense ground truth intrinsic images, e.g., MPI-Sintel and MIT (*bottom*). Note that for the IIW data (*top*), only an albedo image is directly estimated; the shading is then computed via (1).

## 3. Shared Network Structures

Our proposed framework is composed of three central functional components that are largely shared across different dataset types: (i) a direct intrinsic image estimation network (Direct Intrinsic Net), (ii) a sparse guidance map prediction (Guidance Network), and (iii) a reflectance image flattening module (Domain Filter). Figure 1 displays their arrangement, while details are contained below.

### 3.1. Direct Intrinsic Network

Given an input image, an initial coarse estimate $R'$ of the dense intrinsic image decomposition is produced via a 26-layer fully convolutional neural network. The front 3 convolution layers extract a number of feature maps and downscale the resolution to half the input image. The intermediate feature descriptors so-obtained are then fed through the middle 20 dilated convolutional layers, which are reorganized into 10 residual blocks to accelerate network convergence. The output from residual blocks are finally reconstructed to the required intrinsic images by the last 3 convolutional layers. Except for the final convolution, all the other layers share the kernel size (3×3) and the channels for the feature maps (32), and all are followed by batch normalization (BN) and ReLU layers.

---

[1] Note also that we requested this data during the preparation of our work; however, we were informed by the authors of [21] that it was not available for distribution.

While this basic structure is inherited for all experiments, minor customizations must be introduced to accommodate the diversity of training data formats, labeling, and size. In particular, for IIW data where labeling is restricted to sparse pairwise comparisons of relative reflectance, we only require that the Direct Intrinsic Network produce a scalar albedo intensity $r$ for every image pixel. Note however that if we adopt the common assumption that the scene lighting is achromatic as is commonly done for IIW data [2, 3, 19, 24, 25], then $r$ can be expanded to the full albedo $R'$ and shading $S$ layers across all 3 color channels using the differentiable transform

$$R'_i = \frac{r_i}{\frac{1}{3}\sum_c(I_i^c)} \cdot I_i, \quad S_i = \frac{\frac{1}{3}\sum_c(I_i^c)}{r_i} \cdot [1,1,1], \quad (2)$$

where $i$ denotes the pixel location and $c$ is the RGB color index. Hence a simple reconstruction layer can easily produce a full intrinsic decomposition as required by later modules, even if for present purposes here we only output a scalar greyscale reflectance map.

In contrast, for datasets like MIT and MPI-Sintel where dense albedo and shading labels are provided and achromatic lighting assumptions do not strictly hold (e.g., the shading image can be colorful per the generative process), it is more suitable for the Direct Intrinsic Network to separately output full albedo and shading layers. Therefore, the basic network structure described above is split into two branches from within the intermediate residual blocks, one for albedo and another for shading. Furthermore, to achieve a better performance using these dense datasets, we expand the depth to 42 convolution layers, and channels for the feature maps to 64.

### 3.2. Guidance Network

The images generated by the Direct Intrinsic Network described above already demonstrate excellent numerical performance. But we nonetheless still observe that the direct estimation of intrinsic images using parameterized convolutions does not always preserve the flattening effects exhibited by natural reflectance images. To tackle this issue, previous approaches have applied post-processing via either a separate optimization step [24, 25] or various filtering operations [19], all of which rely on strong priors and/or additional inputs to generate realistic piecewise constant effects at a high computational cost. Instead, to obviate the need for any expensive post-processing, we leverage a cheap domain filter guided by a learned edge map that highlights key sparse structure indigenous to albedo images.

Given a guidance image $G$ with salient structural information pertaining to $R$ (more on how $G$ is chosen in Section 4), we compute a scalar edge map via

$$E_i(G) = \sum_{j \in \mathcal{N}_2(i)} |\sum_c (G_i^c - G_j^c)|, \quad (3)$$

where $E(G)$ represents the extracted sparse structure of the guided image and $\mathcal{N}_2(i)$ indicates the surrounding points within a 2-pixel distance from point $i$. The output edge map is greyscale and its intensity demonstrates how salient the color transition is at each point.

Our Guidance Network learns a mapping from $I$ to $E(G)$ via a similar network structure as the Direct Intrinsic Network from above. It consists of 18 convolutional layers with 64 feature maps (except for the last one), and we also adopt dilated convolution for the middle residual blocks. Note that the Guidance Network is unchanged for all datasets. Additionally, we have observed that the computed edge map of the guidance image is usually a simplified version of the one computed from the original input image $I$ (i.e., since the guidance image should contain fewer spurious details). Therefore, we feed both the original input image and its associated input edge map $E(I)$ computed via (3) into the Guidance Network to predict the required salient edge guidance map, which we denote $E'$.

### 3.3. Domain Filter

To generate a realistically flattened albedo image, we adopt a guided, edge-preserving domain filter that requires two inputs: the reflectance image $R'$ as produced by our Direct Intrinsic Network, and a scalar guidance map $E'$ as computed by our Guidance Network. The Domain Filter admits an efficient implementation via separable 1D recursive filtering layers applied across rows and columns in an image, which means performing a horizontal pass along each image row, and a vertical pass along each image column iteratively. For an input 1D signal $X$, the filtered output signal $Y$ can be defined on the transformed domain of guidance map $E'$ using

$$Y_i = (1 - g_i)X_i + g_i Y_{i-1}, \quad (4)$$

where $g$ is a function of $E'$ obtained via the method from [13]. In this context, $g_i$ determines the amount of diffusion by controlling the relative contribution of the raw input signal $X_i$ to the filtered signal value at the previous position $Y_{i-1}$ (the 2D case is similar, where $X$ and $Y$ correspond with the reflectance image before and after filtering). The cumulative effect is that if the learned guidance map is large at point $i$, which means there is a strong color transition there, the filtered reflectance at point $i - 1$ will not be propagated to the point $i$. Otherwise, point $i$ will be flattened or averaged with the value at point $i - 1$. Note that similar recursive 1D filtering has been effectively applied to image smoothing [13, 17] and semantic segmentation [9], which are highly-related computer vision applications.

## 4. Flexibly Supervised Loss Layers

This section discusses the flexibly supervised loss layers (see Figure 1) that can be customized to the distinct forms of available ground truth labels. We differentiate two primary categories of loss layers, one for handling pairwise comparison data of albedo intensities, the other for handling dense maps of full albedo and shading decompositions.

### 4.1. Pairwise Comparison Data

We begin with the pairwise relative reflectance judgements as found in the IIW dataset. As no dense ground truth labels are available, [2] introduced the weighted human disagreement rate (WHDR) as the error metric. For the $k$-th pair of connected points denoted $\{k_1, k_2\}$, a human judgement $J_k \in \{1, 2, E\}$ is issued that indicates if point $k_1$ is either *darker* than ($J_k = 1$), *lighter* than ($J_k = 2$), or *equal* to ($J_k = E$) the reflectance of point $k_2$. Given the pixel-wise mean of a predicted albedo image over RGB channels $\overline{R}$, a classification of reflectance pairs can then be calculated as

$$\hat{J}_\delta(\overline{R}_{k_1}, \overline{R}_{k_2}) = \begin{cases} 1 & \text{if } \overline{R}_{k_2}/\overline{R}_{k_1} > 1 + \delta, \\ 2 & \text{if } \overline{R}_{k_1}/\overline{R}_{k_2} > 1 + \delta, \\ E & \text{otherwise.} \end{cases} \quad (5)$$

where $\delta$ quantifies a significant threshold for the relative difference between two surface reflectances. The WHDR measures the percent of human judgements $J_k$ that an algorithm estimate $\hat{J}_\delta(\overline{R}_{k_1}, \overline{R}_{k_2})$ disagrees with, weighted by a separate confidence score $w_k$ of each judgement. This metric is naturally converted to a form of modified hinge loss that can be conceptually evaluated at every possible pair of points across a dense, trainable albedo image estimate. But for those pairs of points for which no human label is available, we can implicitly assume that $w_k = 0$ (i.e., zero confidence). The albedo image output from the Domain Filter can then be supervised as

$$\mathcal{L}_{df} = \sum_{k \in \varepsilon} w_k \cdot \mu(J_k, \overline{R}_{k_1}, \overline{R}_{k_2}, \delta, \xi), \quad (6)$$

where $\varepsilon$ indicates the set of all the connected points within the image. The function $\mu(J_k, \overline{R}_{k_1}, \overline{R}_{k_2}, \delta, \xi)$ behaves like a standard, SVM hinge loss term with respect to the ratio $\overline{R}_{k_1}/\overline{R}_{k_2}$ when $J_k \in \{1, 2\}$, or an analogous $\epsilon$-insensitive regression loss when $J_k = E$ (see the supplementary file for the exact form of $\mu$). The additional hyper-parameter $\xi$ can be viewed as controlling the margin between neighboring classes as described in [19]. Similar hinge loss functions have also been incorporated into previous intrinsic image decomposition work [19, 24]. For present purposes here, this loss is appealing, since the input image just needs a single forward pass through the network, and the predicted reflectance output can then be used to compute the error metric summed over all the connected points, which cannot be achieved by widely used softmax loss.

Beyond this supervision at the output of our pipeline, we also provide intermediate supervision both to the greyscale albedo intensity $r$ produced by the Direct Intrinsic Network (which per our modeling assumptions captures all degrees of freedom in the initial albedo estimate $R'$ as computed via (2)), and the salient edge map $E'$ produced by the Guidance Network. Regarding the former, the relevant supervision layer is given by

$$\mathcal{L}_{di} = \sum_{k \in \varepsilon} w_k \cdot \mu(J_k, r_{k_1}, r_{k_2}, \delta, \xi). \quad (7)$$

In contrast, supervision on the predicted guidance map is a simple mean squared error,

$$\mathcal{L}_g = ||E' - E(G^*)||_2^2 \quad (8)$$

where $G^*$ denotes a ground truth guidance image. For IIW we have no access to the true albedo images, making a dense optimal selection for $G^*$ infeasible. However, if we assume that the significant edges from the raw image $I$ predominately originate from the implicit albedo component, then we may treat salient edges extracted from $I$ as a rough proxy for salient edges extracted from the unknown optimal $R^*$.

To this end, we compute $G^* = f(I)$, such that $E(G^*) = E(f(I)) \approx E(R^*)$ as the ground truth guidance image, where $f$ is the flattening image filter from [3], which produces piecewise, salient edge-aware effects. The inclusion of this loss term, as well as the subsequent guided Domain Filter, helps to stabilize the network performance when extrapolating to unsupervised image locations underlying the predicted dense map, and leads to more visually realistic, flattened reflectance images. The overall loss then becomes the weighted combination of energy functions given by

$$\mathcal{L} = \mathcal{L}_{di} + \lambda_1 \mathcal{L}_g + \lambda_2 \mathcal{L}_{df}, \quad (9)$$

where $\lambda_1 = 0.35$ and $\lambda_2 = 0.1$ for all experiments.

### 4.2. Densely Labeled Data

When dense ground truth intrinsic images $R^*$ and $S^*$ are available as in MIT and MPI-Sintel datasets, we directly utilize the mean squared error as the supervision layer for all outputs. For the output of the Domain Filter that flattens the albedo component, we therefore adopt the loss

$$\mathcal{L}_{df} = ||R - R^*||_2^2. \quad (10)$$

In contrast, because we have access to the full ground truth for both albedo and shading layers, and given that Equation 1 is only an approximation (meaning both of these components can actually contribute non-trivial information[2]), for the output of the Direct Intrinsic Network we supervise both $R'$ and $S$. Additionally, to help preserve the details of intrinsic images, the image gradients in the x and y directions are also supervised, producing the aggregate intermediate loss

$$\mathcal{L}_{di} = \lambda_2(||R' - R^*||_2^2 + ||S - S^*||_2^2) \\ + \lambda_1(||\nabla_x R' - \nabla_x R^*||_2^2 + ||\nabla_y R' - \nabla_y R^*||_2^2 \\ + ||\nabla_x S - \nabla_x S^*||_2^2 + ||\nabla_y S - \nabla_y S^*||_2^2). \quad (11)$$

Finally, the loss for the Guidance Network is exactly the same as in (8), only now we define $G^* = R^*$ for the guidance filter ground truth. We then jointly train the whole network using

$$\mathcal{L} = \mathcal{L}_{di} + \lambda_1 \mathcal{L}_g + \lambda_2 \mathcal{L}_{df}, \quad (12)$$

with $\lambda_1 = 0.35$ and $\lambda_2 = 0.2$.

---

[2]This is especially true given that MPI-Sintel data contains some defective pixels and the MIT data has a mask.

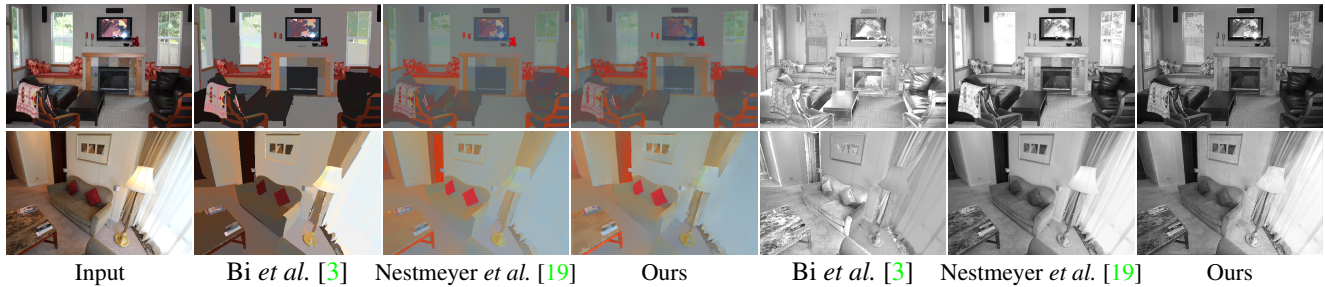

| | | | | | | |
|---|---|---|---|---|---|---|
| Input | Bi *et al.* [3] | Nestmeyer *et al.* [19] | Ours | Bi *et al.* [3] | Nestmeyer *et al.* [19] | Ours |

Figure 2. Qualitative comparison on the IIW benchmark. The second through forth columns represent albedo components, and the fifth through seventh columns are the corresponding shading layers.

| Methods | WHDR (mean) |
|---|---|
| Baseline (const shading) | 51.37 |
| Baseline (const reflectance) | 36.54 |
| Shen *et al.* 2011 [20] | 36.90 |
| Retinex (color) [14] | 26.89 |
| Retinex (gray) [14] | 26.84 |
| Garces *et al.* 2012 [12] | 25.46 |
| Zhao *et al.* 2012 [23] | 23.20 |
| $L_1$ flattening [3] | 20.94 |
| Bell *et al.* 2014 [2] | 20.64 |
| Zhou *et al.* 2015 [24] | 19.95 |
| Nestmeyer *et al.* 2017 (CNN) [19] | 19.49 |
| Zoran *et al.* 2015* [25] | 17.85 |
| Nestmeyer *et al.* 2017 [19] | 17.69 |
| Bi *et al.* 2015 [3] | 17.67 |
| Ours w/o D-Filter | 15.40 |
| Ours w/o joint training | 14.52 |
| Ours | **14.45** |

Table 1. Quantitative results on the IIW benchmark. All the results are evaluated on the test split of [18], except for the one marked with * which is evaluated on their own test split and is not directly comparable with other methods.

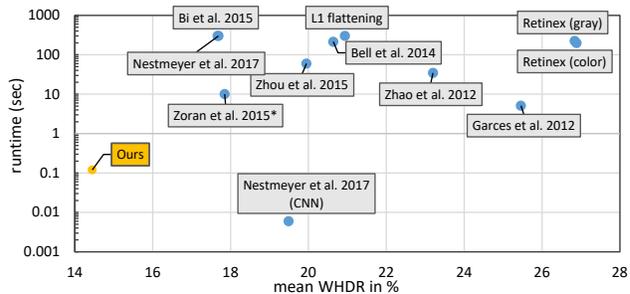

Figure 3. WHDR against runtime plot. The WHDRs are consistent with Table 1, while the running times of the previous methods are collected from [19]. Our algorithm achieves the best performance on WHDR and takes less than 100ms for evaluation.

## 5. Experimental Results

### 5.1. Sparse Pairwise Supervision via IIW Data

**Datasets:** The Intrinsic Images in the Wild (IIW) benchmark [2] contains 5,230 real images of mostly indoor scenes, combined with a total of 872,161 human judgements regarding the relative reflectance between pairs of points sparsely selected throughout the images. Consistent with many prior works [18, 24, 19], we split the IIW dataset by placing the first of every five consecutive images sorted by the image ID into the test set while the others are used for training. For quantifying the quality of reconstructed albedo images, we employ the WHDR from [2] and as described in Section 4.

**Comparison:** Table 1 presents the numerical results, where our full pipeline achieves the best performance (mean WHDR 14.45), which is significantly better than the second best one [3] (mean WHDR 17.67). To further deconstruct the effectiveness of our developed framework, we also include an ablation study in the bottom of Table 1. Here we observe that the domain filter and learned guidance map do significantly improve the performance. Moreover, we find that our algorithm also benefits from joint training the entire pipeline. In terms of computational complexity, Figure 3 displays runtime comparisons plotted against the WHDR performance metric across a wide array of competing methods. Note that even while obtaining the highest accuracy score, our system is still faster than most others.

Finally, some representative visual examples are presented in Figure 2, which illustrate dense extrapolated decompositions across entire images. The results from [3] show some abrupt color transitions along the front and side of the couch that, at least by visual inspection, presumably should have the same albedo. Likewise, the reflectance estimate from [19] contains spurious noise in many places, and is not nearly as flattened as ours.

### 5.2. Dense Supervision via MPI-Sintel and MIT Data

**Datasets:** We follow two recent state-of-the-art deep learning based methods [15, 22] and evaluate our algorithm on the MPI-Sintel dataset [4] that facilitates scene-level quantitative comparisons. This dataset consists of 890 images from 18 scenes with 50 frames each (except for one that contains 40 images). Due to limited images in this dataset, we randomly crop 10 different patches of size 300×300 from one image to generate 8900 patches. Like [22], we use two-fold cross validation to obtain all 890 test results with two trained models. We evaluate our results on both a *scene split*, where half the scenes are used for training and the other half for testing, and an *image split*, where all 890 images are randomly separated into two parts.

|  |  | MSE | | | LMSE | | | DSSIM | | |
|---|---|---|---|---|---|---|---|---|---|---|
|  |  | albedo | shading | average | albedo | shading | average | albedo | shading | average |
| *image split* | Retinex [14] | 0.0606 | 0.0727 | 0.0667 | 0.0366 | 0.0419 | 0.0393 | 0.2270 | 0.2400 | 0.2335 |
|  | Barron *et al.* [1] | 0.0420 | 0.0436 | 0.0428 | 0.0298 | 0.0264 | 0.0281 | 0.2100 | 0.2060 | 0.2080 |
|  | Chen *et al.* [10] | 0.0307 | 0.0277 | 0.0292 | 0.0185 | 0.0190 | 0.0188 | 0.1960 | 0.1650 | 0.1805 |
|  | MSCR [22] | 0.0100 | 0.0092 | 0.0096 | 0.0083 | 0.0085 | 0.0084 | 0.2014 | 0.1505 | 0.1760 |
|  | Ours | **0.0069** | **0.0059** | **0.0064** | **0.0044** | **0.0042** | **0.0043** | **0.1194** | **0.0822** | **0.1008** |
| *scene split* | MSCR [22] | 0.0190 | 0.0213 | 0.0201 | 0.0129 | 0.0141 | 0.0135 | 0.2056 | 0.1596 | 0.1826 |
|  | Ours | **0.0189** | **0.0171** | **0.0180** | **0.0122** | **0.0117** | **0.0119** | **0.1645** | **0.1450** | **0.1547** |

Table 2. Quantitative comparison on the main MPI-Sintel benchmark. We evaluate our results using both scene and image splits across three standard error rates of intrinsic images on the main MPI-Sintel dataset.

|  |  | MSE | | | LMSE | | | DSSIM | | |
|---|---|---|---|---|---|---|---|---|---|---|
|  |  | albedo | shading | average | albedo | shading | average | albedo | shading | average |
| *image split* | JCNF [15] | 0.0070 | 0.0090 | 0.0080 | 0.0060 | 0.0070 | 0.0065 | **0.0920** | 0.1010 | 0.0970 |
|  | Ours | **0.0040** | **0.0052** | **0.0046** | **0.0030** | **0.0040** | **0.0035** | 0.1081 | **0.0815** | **0.0948** |

Table 3. Quantitative comparison on the auxilliary MPI-Sintel benchmark. Note that JCNF [15] is only trained and tested on the *image split* of MPI-Sintel dataset; hence our exclusion of the scene split here.

While investigating the MPI-Sintel dataset online, we noticed that there are actually two sources for the input and albedo images. The first one is obtainable by emailing the authors of [4] directly, while the second one can be partially downloaded from their official web page (but also requires emailing to obtain full ground-truth). We refer to them as main and auxiliary MPI-Sintel dataset separately based on their popularity among the research community.

Finally, to test performance on real images where scene-level ground-truth is unavailable, we also use the 220 images in the MIT intrinsic dataset [14] as in [22]. This data contains only 20 different objects, each of which has 11 images. To compare with previous methods, we train our model using 10 objects via the split from [1], and evaluate the results on images from the remaining objects.

**Comparison:** As shown in Table 2 and 3, our model achieves the best result for most columns on the MPI-Sintel data. Note the other methods [15, 22] also benefit from training on additional training data. We show a group of qualitative results trained on the more difficult *scene split* in Figure 5. It can be seen that our framework produces sharper and high-quality results.

|  | MSE | | | LMSE |
|---|---|---|---|---|
|  | albedo | shading | average | total |
| Barron *et al.* [1] | 0.0064 | 0.0098 | 0.0081 | 0.0125 |
| Zhou *et al.* [24] | 0.0252 | 0.0229 | 0.0240 | 0.0319 |
| Shi *et al.*[21] | 0.0216 | 0.0135 | 0.0175 | 0.0271 |
| MSCR [22] | 0.0207 | 0.0124 | 0.0165 | 0.0239 |
| Ours | **0.0134** | **0.0089** | **0.0111** | **0.0203** |

Table 4. Results on the MIT data. Performance of various methods on Barron *et al.*'s test set [1]. LMSE is computed using an error metric specifically designed for this data [14]. Note also that Barron *et al.*'s approach [1] relies on specialized priors and masked objects particular to this dataset.

Next, Table 4 presents the relative performance on the MIT intrinsic data. We observe that our approach is also the best compared with the other deep networks [21, 22, 24] even though [21, 22] utilize additional training data. Note that [1] uses a number of specialized priors appropriate for this simplified object-level data, while end-to-end CNN approaches like ours and [22] have less advantage here due to limited training data (110 images). Moreover, [1] is not competitive on other more complex, scene-level data types as shown in Table 2. In Figure 4, our predicted images are also sharper and more accurate than the other deep methods.

### 5.3. Joint Supervision via Multiple Data Sources

Simultaneously training on multiple datasets is a natural consideration given the generic, modular nature of our pipeline. To briefly examine this hypothesis, we jointly trained our model on both IIW and MPI datasets, with shared parameters for the Direct Intrinsic and Guidance networks (note that although there are several compelling ways to merge objectives, we omitted supervision on the shading component from MPI data for simplicity here). Moreover, to balance gradients from the two quite different loss layers, modified hinge loss for IIW and MSE for MPI, we scale the former as two times the latter.

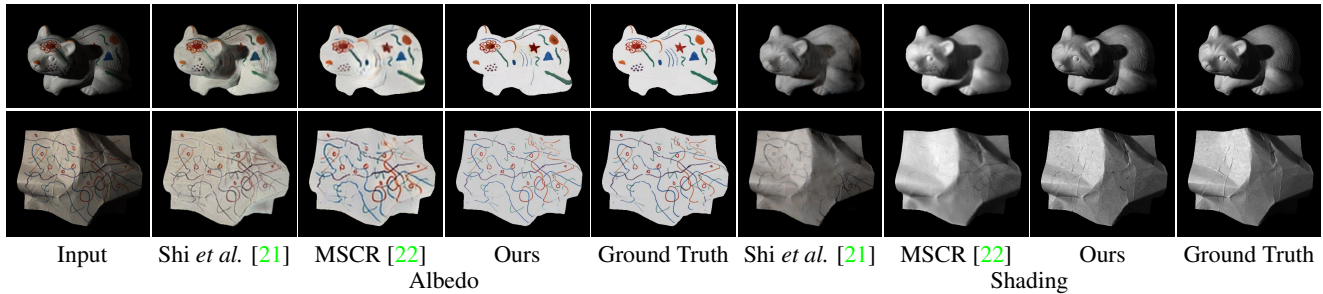

Figure 4. Qualitative comparison on the MIT intrinsic image benchmark. Compared with Shi *et al.* [21] and MSCR [22] on Barron *et al.*'s test split, our algorithm achieves better results.

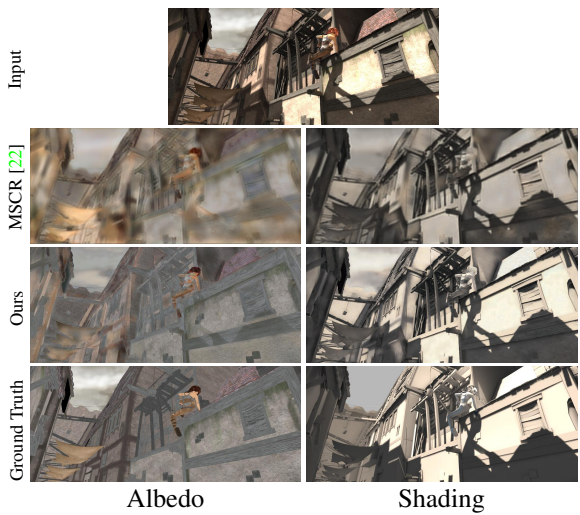

Figure 5. Qualitative comparison on the main MPI-Sintel benchmark. The visual results are evaluated on the model trained on the more difficult *scene split*.

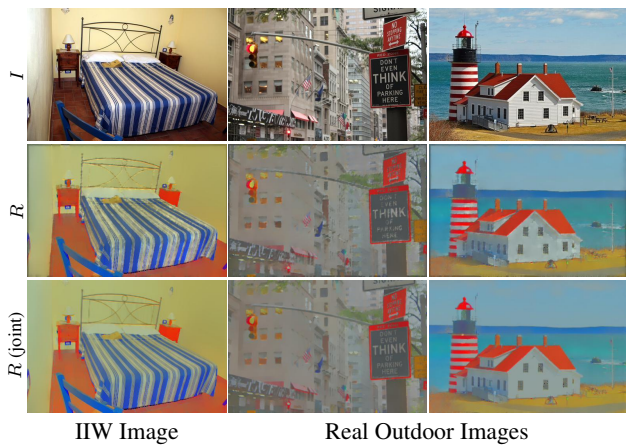

Figure 6. Reflectance estimates generated by our deep network when trained with only IIW data, or jointly trained with IIW and MPI data. The joint training yields smoother and much more realistic results on completely new, real-world outdoor scene images that are not a part of either dataset. **Zoom to see details**.

Experimentally, we obtained a mean WHDR of 15.80 on IIW, better than all previous methods but not quite as good as our previous result when trained on IIW only. This is not surprising since the dense, synthesized MPI data is unlikely to closely reflect real-world images and IIW pairwise comparisons. But crucially, MPI data can still provide useful regularization/smoothing of real-world image structures, even though this benefit may occur away from sparsely-labeled points interior to different surface materials and hence, contributes no quantitatively measurable value per the WHDR criterion.

Figure 6 supports this conclusion, where our jointly trained model is applied to three real-world images, one from IIW, and two from an independent source. Here we observe that complementary supervision does in fact enhance the qualitative performance in new testing environments and our joint model smooths various artifacts.

## 6. Conclusion

In this paper, we solve the intrinsic image decomposition problem using a unified deep architecture that produces state-of-the-art results, with a minimal computational footprint, whether trained on weakly labeled pairwise comparison from IIW data or dense ground truth images from MIT or MPI-Sintel datasets. Our network is end-to-end trainable, requires no expensive post-processing, and is able to generate realistically-flattened dense intrinsic images even on the more challenging IIW dataset. We conjecture that the modular structure we propose will also seamlessly adapt to new sources of labeled data.

**Acknowledgement** We would also like to acknowledge our research grants: National 973 Program (2015CB352501), NSFC-ISF (61561146397) and Shenzhen Innovation Program (JCYJ20150402105524053).

# Revisiting Deep Intrinsic Image Decompositions
## (*Supplementary Material*)

Qingnan Fan[*1]  Jiaolong Yang[2]  Gang Hua[2]  Baoquan Chen[3,4]  David Wipf[2]
[1]Shandong University  [2]Microsoft Research
[3]Shenzhen Research Institute, Shandong University  [4]Peking University
fqnchina@gmail.com, {jiaoyan,ganghua,davidwip}@microsoft.com, baoquan@pku.edu.cn

# 1  Outline

This document contains several technical details and experiments that could not be included in the main paper due to space constraints. The remaining content is organized as follows.

- **Section 2:** More details of joint training with multiple datasets.
- **Section 3:** Details of the hinge loss used for pairwise comparison data in the IIW dataset.
- **Section 4:** Training details for the proposed framework.
- **Section 5:** More qualitative results on the IIW benchmark with intermediate outputs.
- **Section 6:** Test of generalization to outdoor images.
- **Section 7:** More visual results on the MIT intrinsic dataset.
- **Section 8:** Additional details regarding MPI-Sintel data.
- **Section 9:** More visual comparisons with state-of-the-art methods on MPI-Sintel dataset.

# 2  More Implementation Details about Joint Training of Multiple Data Sources

Joint training under this circumstances can be conducted with any combination of the three datasets. Since improving the performance on real images is a more practical and meaningful solution to intrinsic image decompositions, we base our discussion mainly on the IIW dataset and its specific network structure.

Note as the guidance network and domain filter is employed only to further improve the piecewise constant effects while the primary intrinsic results are predicted by the direct intrinsic network, our modifications are mainly based on the DI network, which are detailed below.

(**i**). The image split of the main MPI benchmark is incorporated for joint training, instead of MIT. The MPI dataset contains enough quantity of training data compared to IIW data (4370 v.s. 4184), while MIT only has 110 training images, which is far from comparable to the others. (**ii**). The last layer of DI network with 1-channel output is adaptively changed to 3 channels to be compatible with MPI data. Then the albedo intensity is later calculated as the mean of three output channels. (**iii**). The loss layer for MPI data contains only MSE criterion for the albedo image without gradient supervision and consideration of the shading image. (**iv**). The weight to balance the loss function between IIW and MPI is empirically determined as 1 and 0.5 due to dramatically different back propagated gradients. (**v**). We jointly train the direct intrinsic network with both datasets, while finetuning the whole framework with only the IIW data.

---

[*]This work was done when Qingnan Fan was an intern at MSR.

## 3 Hinge Loss for Pairwise Comparison Data

The classification of reflectance pairs for the definition of WHDR error metric is calculated as

$$\hat{J}_\delta(\overline{R}_{k_1}, \overline{R}_{k_2}) = \begin{cases} 1 & \text{if } \overline{R}_{k_2}/\overline{R}_{k_1} > 1+\delta, \\ 2 & \text{if } \overline{R}_{k_1}/\overline{R}_{k_2} > 1+\delta, \\ E & \text{otherwise.} \end{cases} \quad (1)$$

Therefore, the SVM hinge loss [6] we used for pairwise comparison data can be written as

$$\mu(J_k, \overline{R}_{k_1}, \overline{R}_{k_2}, \delta, \xi) = \begin{cases} max(0, \frac{\overline{R}_{k_1}}{\overline{R}_{k_2}} - \frac{1}{1+\delta+\xi}) & \text{if } J_k = 1, \\ max(0, 1+\delta+\xi - \frac{\overline{R}_{k_1}}{\overline{R}_{k_2}}) & \text{if } J_k = 2, \\ max(0, \begin{cases} \frac{1}{1+\delta-\xi} - \frac{\overline{R}_{k_1}}{\overline{R}_{k_2}}, \\ \frac{\overline{R}_{k_1}}{\overline{R}_{k_2}} - (1+\delta-\xi) \end{cases}) & \text{if } J_k = E. \end{cases} \quad (2)$$

## 4 Training Details

The proposed framework is implemented with Torch. We adopt the Adam solver [5] for optimizing the loss functions. The learning rate is initialized with 0.01, and decreased to 0.001 to generate further improvement. The network parameters are initialized with the method described in [3]. The training samples are randomly shuffled at the beginning of each epoch. We first train the direct intrinsic network and guidance network separately and in parallel, then jointly train (finetune) the entire network. The whole training procedure takes up to 12 hours on an NVIDIA Tesla P100 GPU.

## 5 More Visual Results on the IIW dataset

Figure 1 shows more visual results on the IIW dataset. The intermediate results are also presented for better understanding of our framework. The guidance network predicts more sparse and salient structures ($E'$) compared to the input ($E(I)$). Guided by the learned edge map ($E'$), the initially predicted reflectance ($R'$) is flattened by domain filter and appears more realistic and piecewise constant. From the final output albedo and shading images, it can be observed that lighting and surface reflectance are well separated.

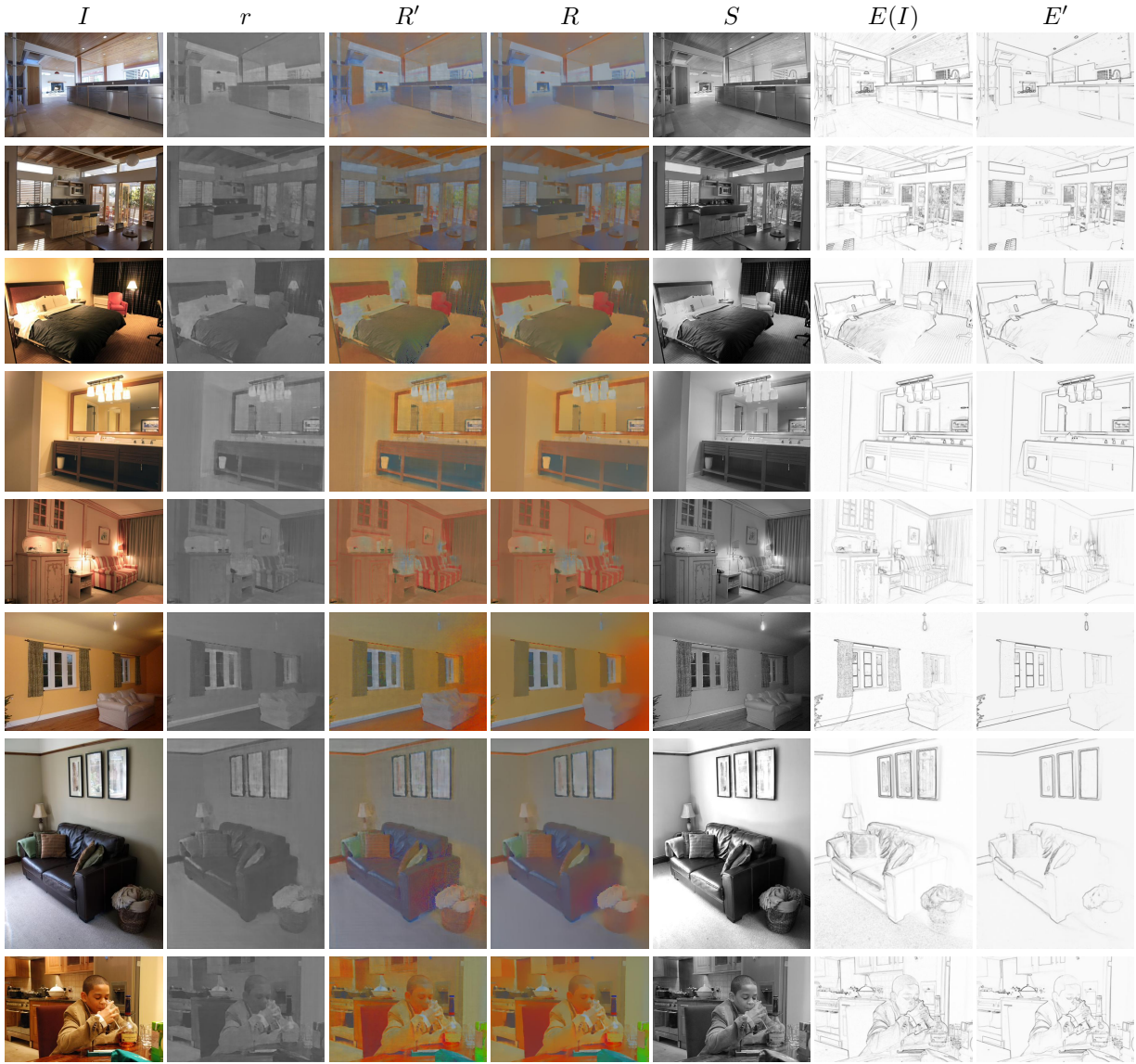

Figure 1: More visual results on the IIW dataset with intermediate outputs. Note the filtered albedo images ($R$) guided by predicted edge map ($E'$) are smoother and more realistic compared to the unfiltered albedo ($R'$). As can be seen, the lighting $S$ is mostly separated from the reflectance $R$.

# 6 Test of Generalization to Outdoor Images

In this section, we evaluate the network trained on IIW dataset, which is composed of mostly indoor images, on outdoor scenes from a completely different, unrelated data source. As shown in Figure 2, the lighting and reflectance are well separated from each other for these outdoor images. For example, the dark shadow under the truck is mostly classified as part of shading layer. These results demonstrate the effectiveness of our framework on various different scenes.

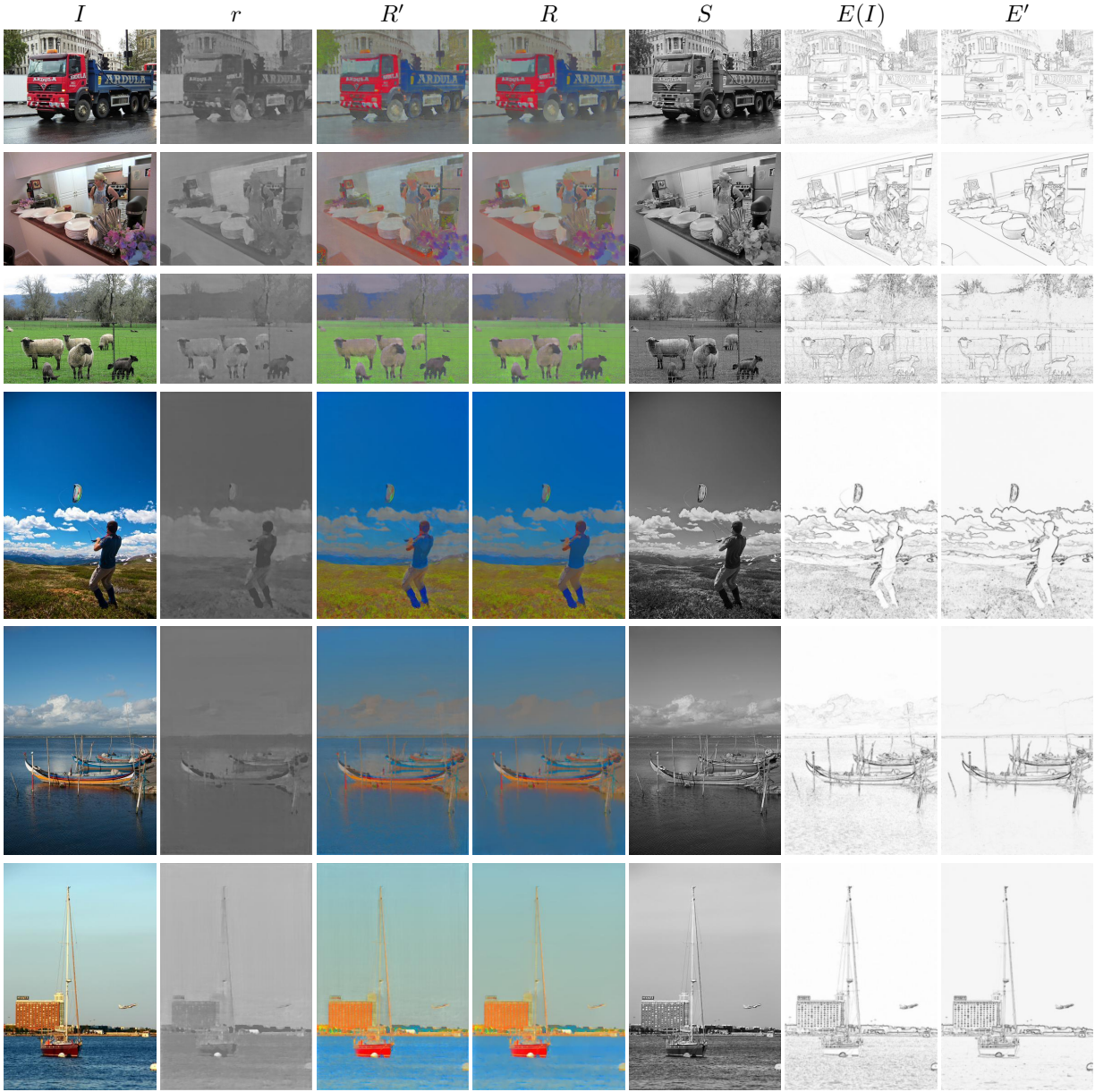

Figure 2: Evaluation of the generalization ability to outdoor scenes of the network trained on the IIW dataset which contains mostly indoor images.

# 7 More Visual Results on the MIT intrinsic dataset

In this section, we show more qualitative results on the MIT intrinsic benchmark. Note our proposed network is only trained on 110 images, and the training and testing splits have no overlap. It can be seen in Figure 3 that the shading is also recovered well in some dark shadow regions such as on the top right of turtle case.

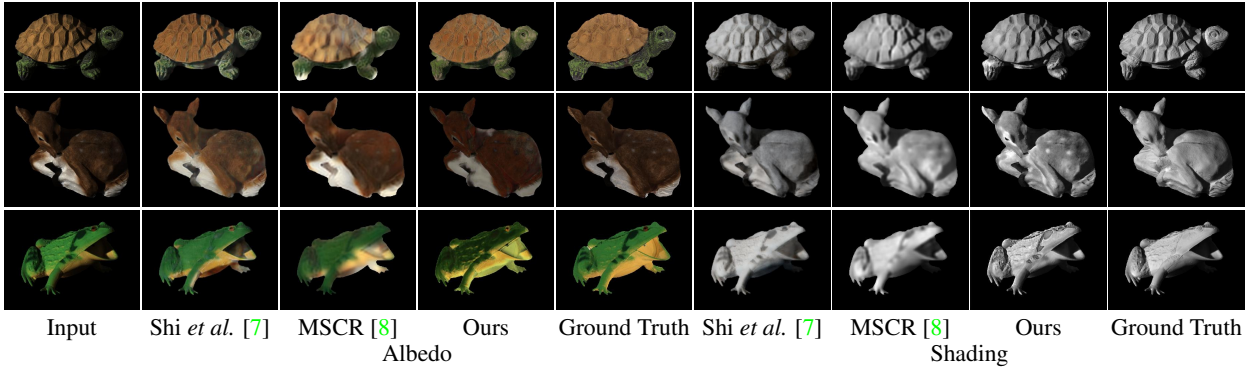

Figure 3: **Qualitative Comparison on MIT Intrinsic Benchmark**. Compared with Shi *et al.* [7] and MSCR [8] on Barron *et al.*'s test split of MIT intrinsic dataset, our algorithm achieves the best/sharpest results.

# 8 Additional Details Regarding MPI-Sintel dataset

Figure 4 compares the images in the *main* and *auxiliary* versions (discussed in the main paper) of the MPI-Sintel benchmark. Note also that the main MPI-Sintel benchmark used by most previous methods contains many defective pixels. We remove the defective images for evaluation on the *image split* following [8], and remove two scenes, "bandage_1" and "shaman_3", for evaluation on the *scene split*. Additionally, to calculate the error on the MIT intrinsic dataset, we use the publicly-available evaluation code of [1] for all methods.

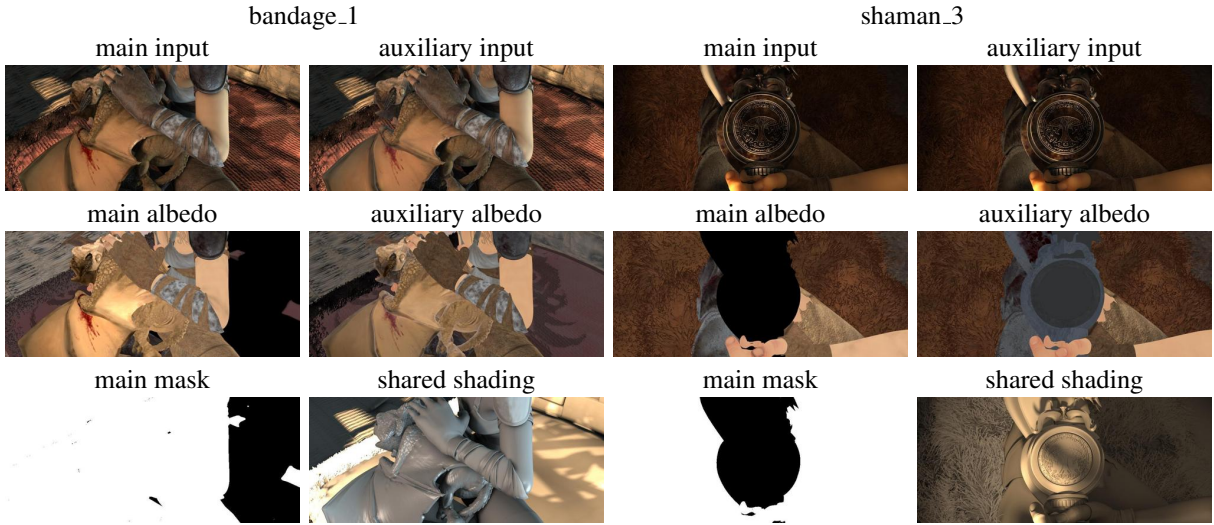

Figure 4: Illustration of differences between the *main* and *auxiliary* versions of the MPI-Sintel benchmark. Note that in the main version, the above albedo images contain many defective pixels due to rendering issues, which is not the case in the auxiliary version. The shading image is shared by the main and auxiliary datasets.

# 9 More Visual Results on MPI-Sintel dataset

This section presents more visual results of our approach compared against existing intrinsic image decomposition methods. Specifically, using the *image split* of the MPI-Sintel benchmark, we provide visual comparisons with several traditional and deep learning based algorithms on the main version of the dataset in Figure 5 and 6. Additionally, we compare our algorithm with [4] on the auxiliary version of the dataset in Figure 7, since [4] is the only previous method trained and tested on this version. On the *scene split* which is much more difficult, we show some qualitative results in Figure 8 and 9.

Our method does not require any additional training data or information such as depth. Compared to the state-of-the-art methods, our results are much sharper, clearer and more piecewise constant on both the *image split* and *scene split* of the main and auxiliary versions of the MPI Sintel benchmark.

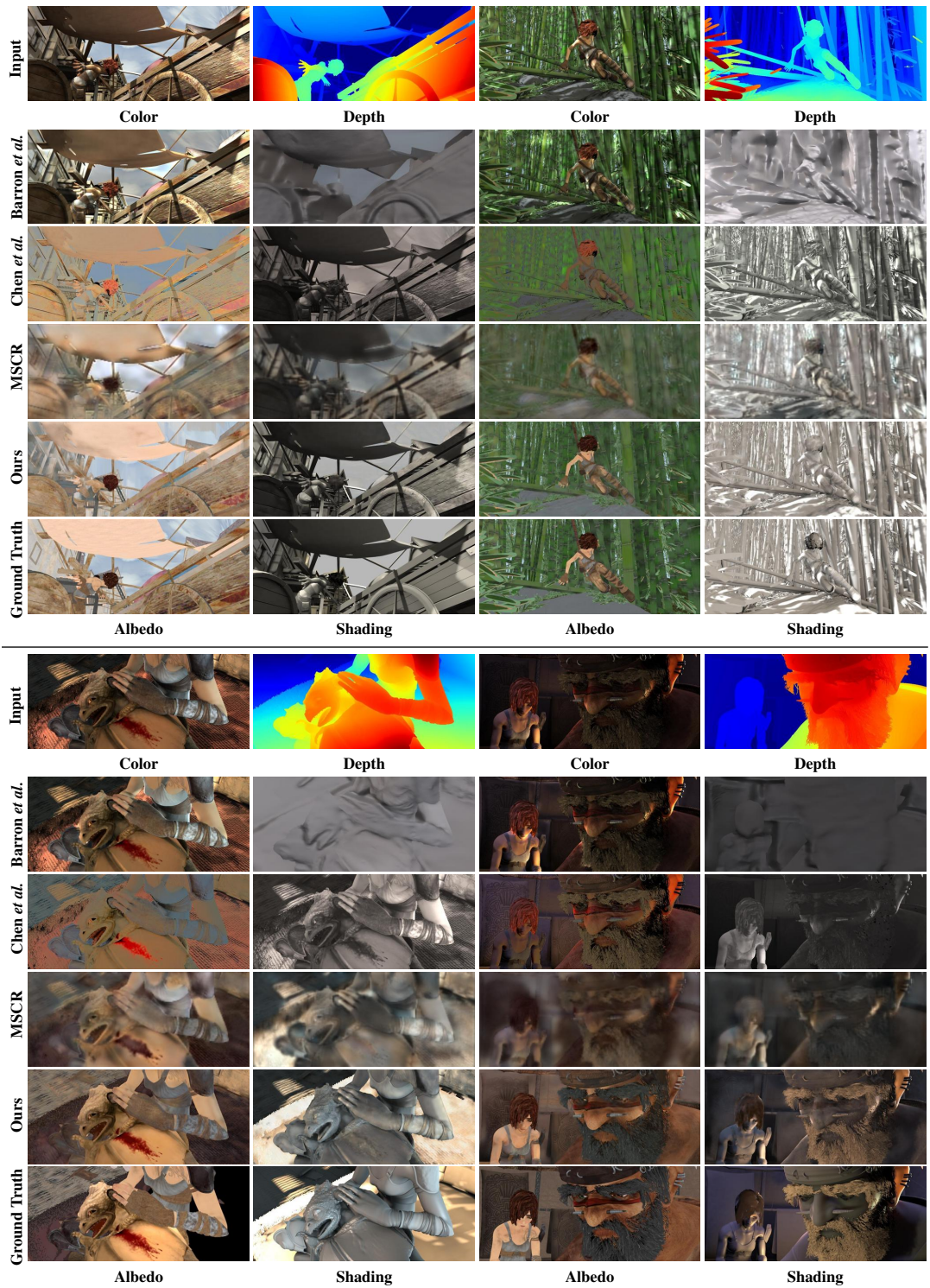

Figure 5: Qualitative comparisons with Barron *et al.* [1], Chen *et al.* [2] and MSCR [8] on the *image split* of the main MPI-Sintel Benchmark.

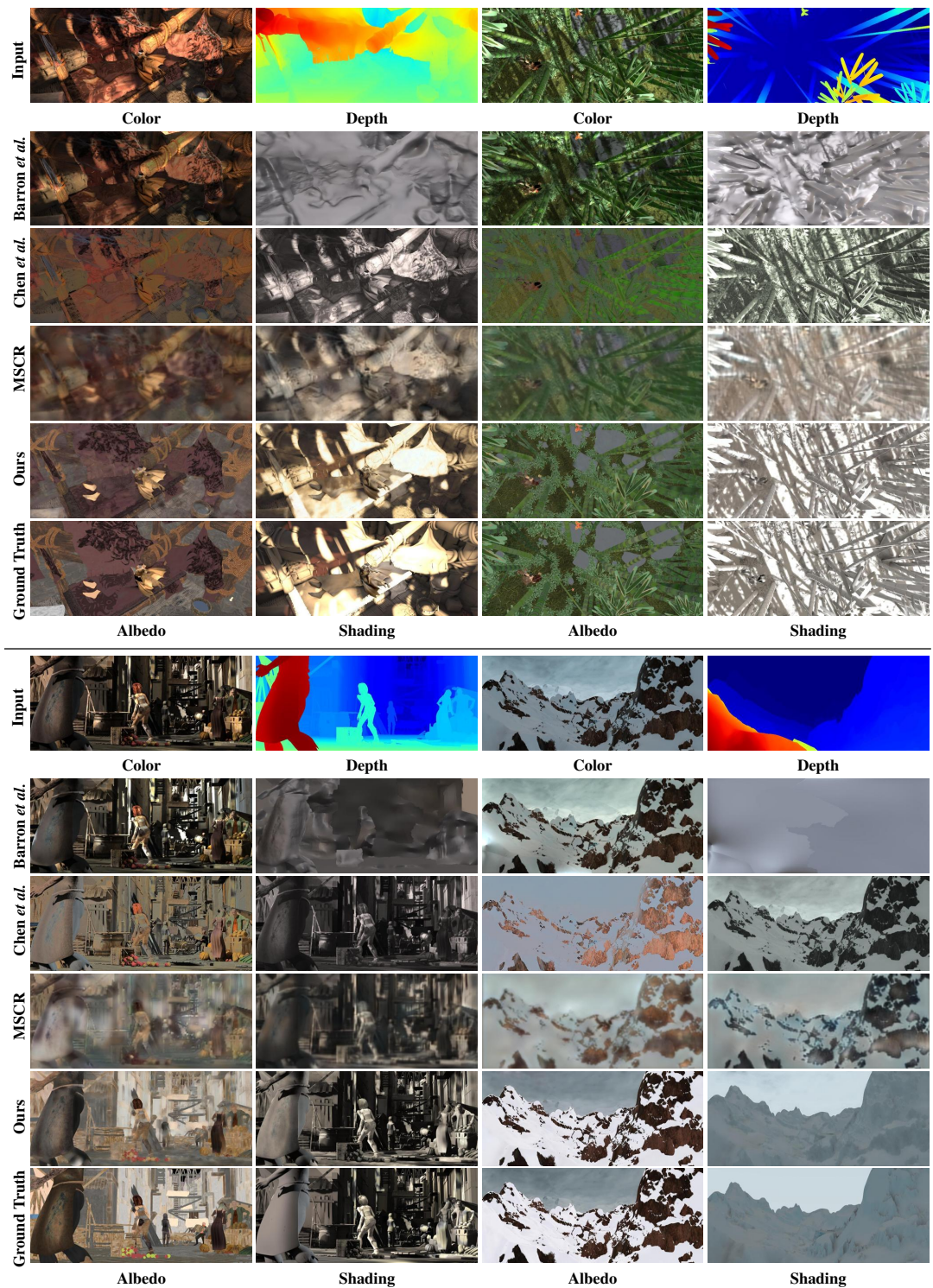

Figure 6: Qualitative comparisons with Barron *et al.* [1], Chen *et al.* [2] and MSCR [8] on the *image split* of the main MPI-Sintel Benchmark.

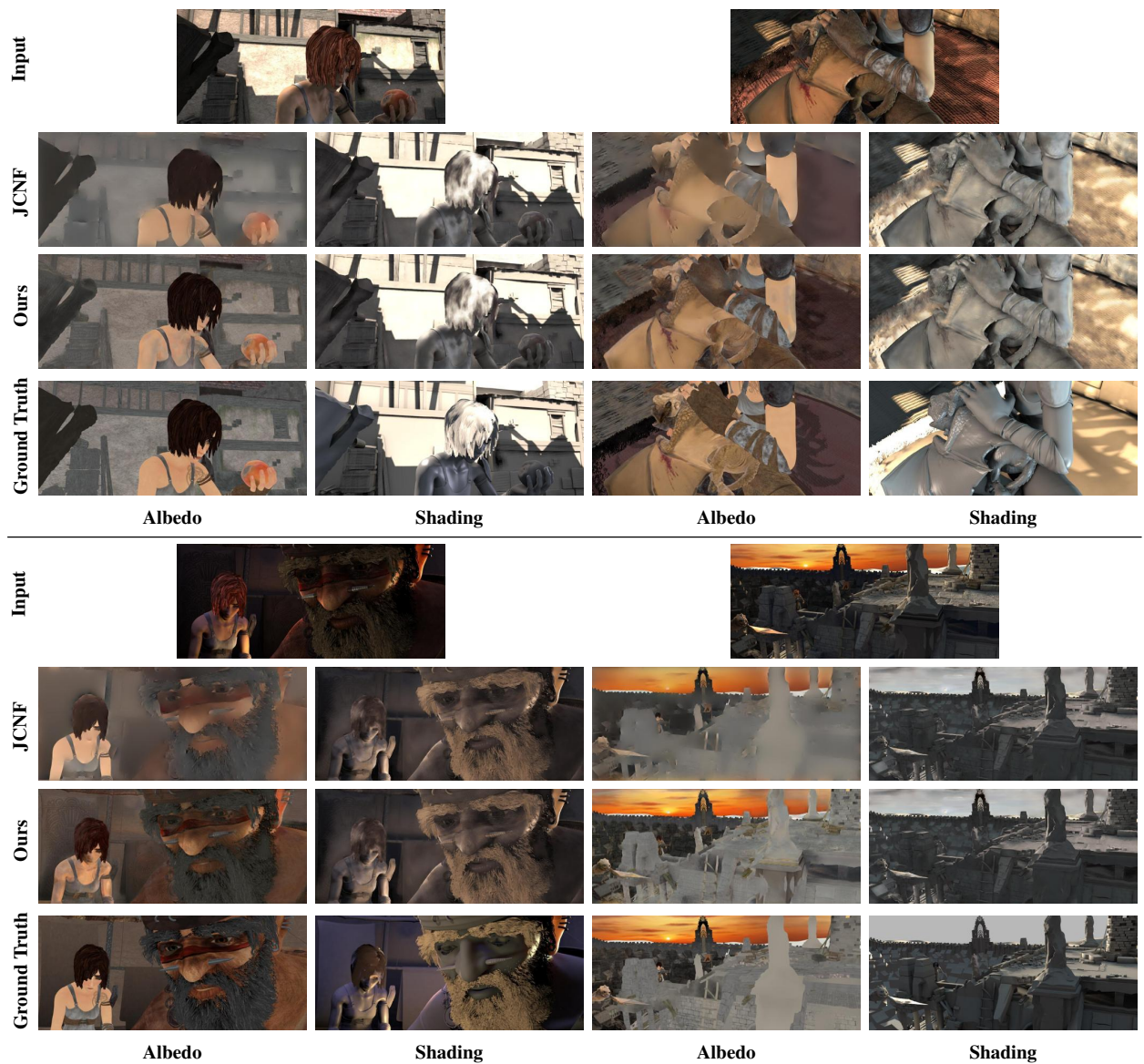

Figure 7: Qualitative comparisons with JCNF [4] on the *image split* of the auxiliary MPI-Sintel Benchmark

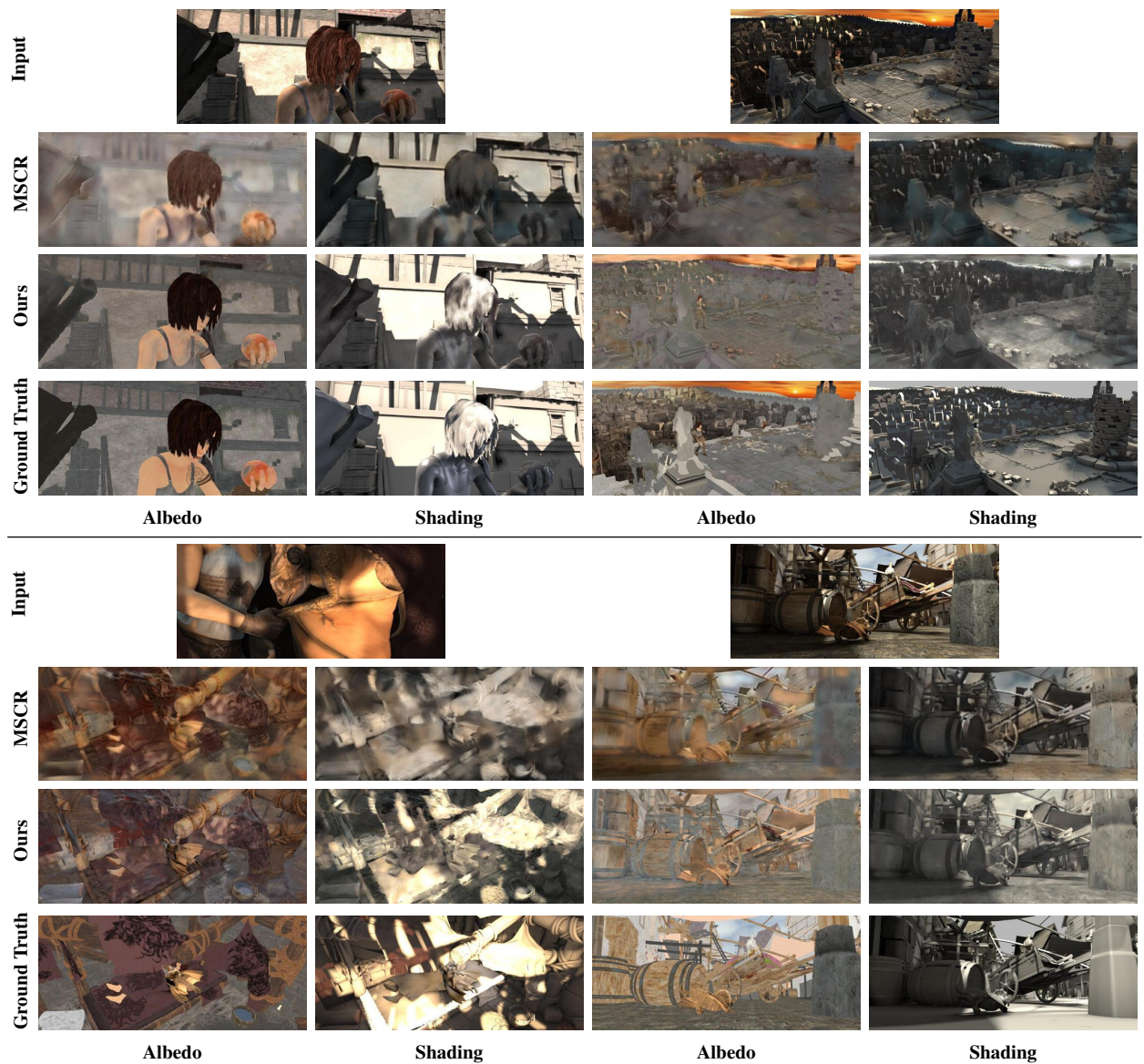

Figure 8: Qualitative comparisons with MSCR [8] on the challenging *scene split* of the MPI-Sintel Benchmark.

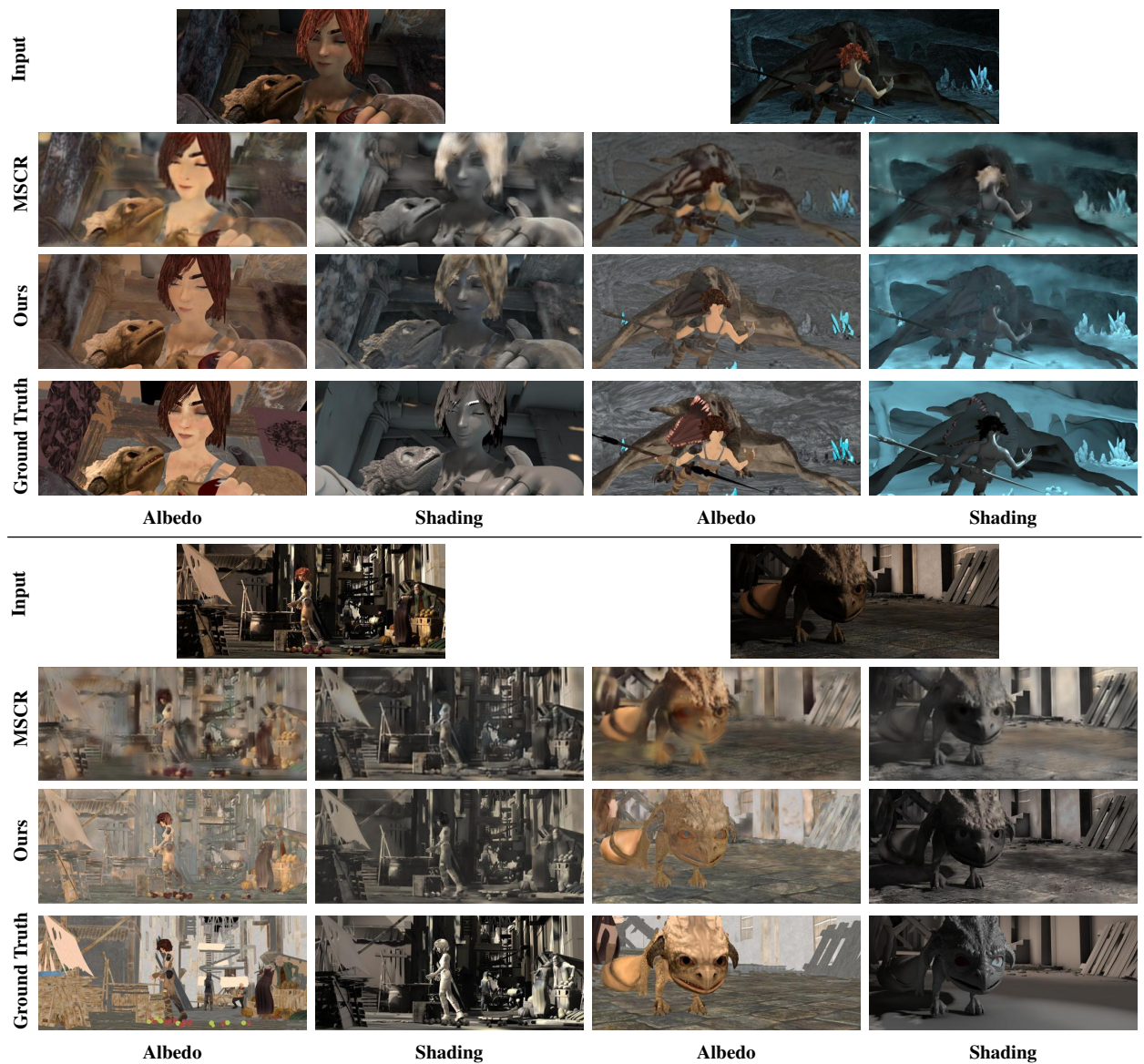

Figure 9: Qualitative comparisons with MSCR [8] on the challenging *scene split* of the MPI-Sintel Benchmark.